\def\year{2020}\relax
\documentclass[letterpaper]{article} 
\usepackage{aaai20}  
\usepackage{times}  
\usepackage{helvet} 
\usepackage{courier}  
\usepackage[hyphens]{url}  
\usepackage{graphicx} 
\usepackage{graphicx}  
\usepackage{amsmath}
\usepackage{amsfonts}
\usepackage{multirow}
\usepackage{subfig}

\newcounter{tightlistcnt}
\newenvironment{tightitemize}{%
	\begin{list}{\textbullet}{\usecounter{tightlistcnt}%
			\topsep 0in
			\partopsep 0in
			\itemsep .2em
			\parsep 0in
			\leftmargin 0em
			\rightmargin 0in
			\listparindent 0em
			\itemindent .75em
			\labelsep .75em
			\labelwidth 0in
		}%
	}{%
	\end{list}%
}

\newenvironment{tightenumerate}{%
	\begin{list}{\arabic{tightlistcnt}.}{\usecounter{tightlistcnt}
			\topsep 0in
			\partopsep 0in
			\itemsep .2em
			\parsep 0in
			\leftmargin 0.0em
			\rightmargin 0in
			\listparindent 0em
			\itemindent .75em
			\labelsep .75em
			\labelwidth 0in
		}%
	}{%
	\end{list}%
}

\title{DialogAct2Vec: Towards End-to-End Dialogue Agent by Multi-Task Representation Learning}
\author{Zhuoxuan Jiang$^1$, Ziming Huang$^2$, Dong Sheng Li$^1$, Xian-Ling Mao$^3$\\
	$^1$IBM Research China, Shanghai, China\\
	$^2$IBM Research China, Beijing, China\\
	$^3$Beijing Institute of Technology, Beijing, China\\
	\{jzxjiang, hzmzi, ldsli\}@cn.ibm.com, maoxl@bit.edu.cn\\
}
 
\begin{document}

\maketitle

\begin{abstract}
In end-to-end dialogue modeling and agent learning, it is important to (1) effectively learn knowledge from data, and (2) fully utilize heterogeneous information, e.g., dialogue act flow and utterances. However, the majority of existing methods cannot simultaneously satisfy the two conditions. For example, rule definition and data labeling during system design take too much manual work, and sequence-to-sequence methods only model one-side utterance information. In this paper, we propose a novel joint end-to-end model by multi-task representation learning, which can capture the knowledge from heterogeneous information through automatically learning knowledgeable low-dimensional embeddings from data, named with DialogAct2Vec. The model requires little manual work for intervention in system design and we find that the multi-task learning can greatly improve the effectiveness of representation learning. Extensive experiments on a public dataset for restaurant reservation show that the proposed method leads to significant improvements against the state-of-the-art baselines on both the act prediction task and utterance prediction task.
\end{abstract}

\section{Introduction}


Task-completion dialogue systems attract extensive attentions from both academic and industrial communities recently, since various domains can benefit from this line of research, such as restaurant reservation~\cite{dstc3}, movie ticket booking~\cite{movie}, client service~\cite{kefu} and travel planning~\cite{hierarchical}. Although most existing commercial dialogue platforms can support to build such an agent, e.g. IBM Watson Assistant\footnote{\url{https://www.ibm.com/cloud/watson-assistant/}}, Google Dialogflow\footnote{\url{https://dialogflow.com/}} and Amazon Alexa\footnote{\url{https://alexa.amazon.com/}}, lots of manual efforts and domain knowledge are required in the process of system design. For example for building an agent, platform users have to define intents, entities, dialogue flows, utterances, etc. It is one of the challenges for NLP and AI communities to learn a dialogue agent from data with as little as possible human intervention.

Conceptually, a dialogue system consists of a pipeline with several modules, such as natural language understanding (NLU), dialogue state tracking (DST), dialogue control (DC, i.e. dialogue policy selection) and natural language generation (NLG). To reduce human efforts, various learning based methods are proposed for training each module. For example, Convolutional Neural Network (CNN) and Long Short-Term Memory (LSTM) are used for intent identification of NLU~\cite{icdm}, Bidirectional Long Short-Term Memory (BiLSTM) and Conditional Random Field (CRF) are used for slot filling of DST~\cite{bilstmcrf1,crflstm}, reinforcement learning methods are suitable for dialogue control~\cite{withhuman}, and sequence-to-sequence framework fits for language generation~\cite{goal1}. Nevertheless, separately learning each module requires a lot of human efforts on labeling data and coordination. Recently, an end-to-end trainable method is proposed~\cite{EACL} to assemble all learned modules in one system, which may suffer from the potential error propagation issue~\cite{errorpro2}.

\begin{figure}
	\centering
	\includegraphics[width=0.49\textwidth]{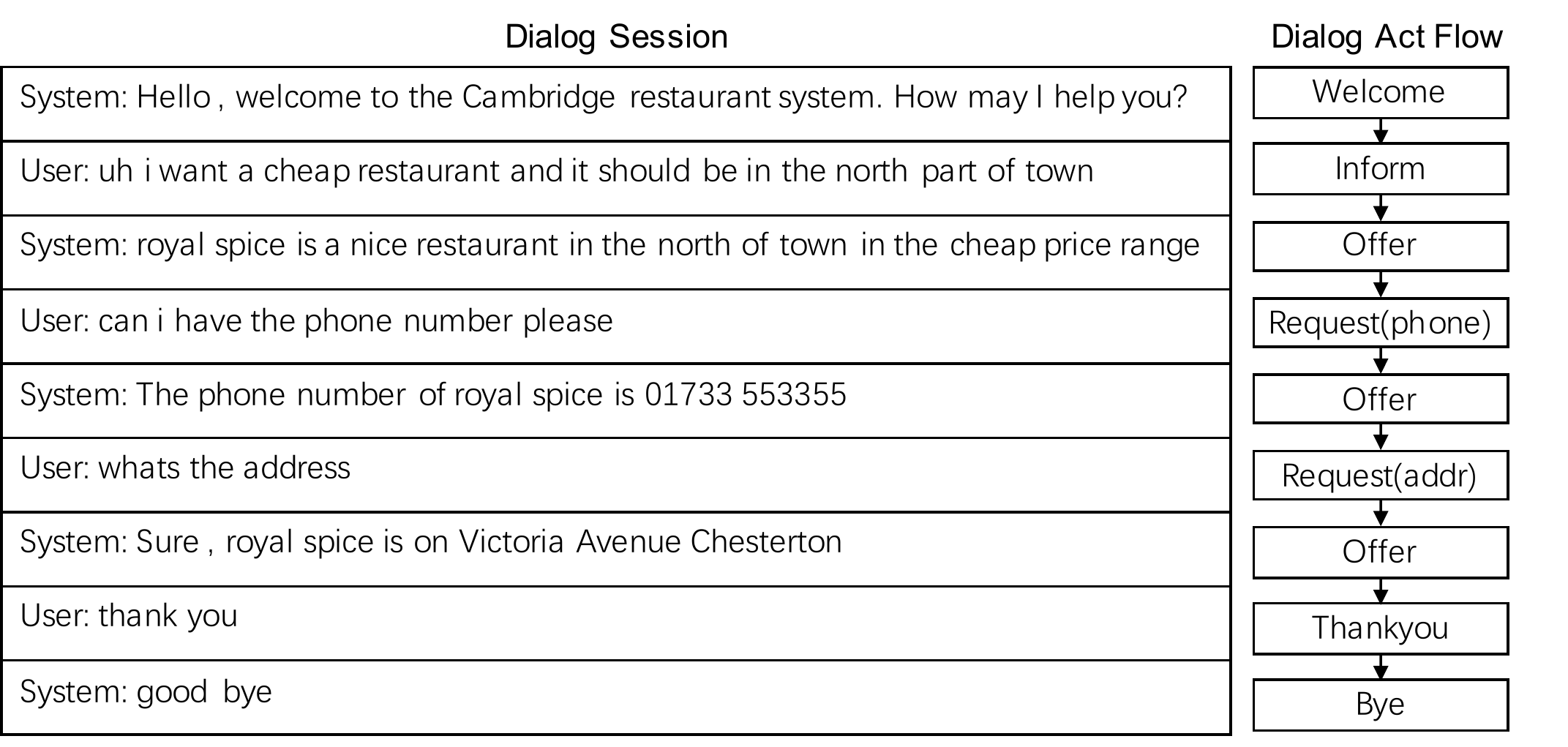}
	\caption{An example of a dialogue session with a dialogue act flow.\label{fig:example}}
\end{figure}

In addition to pipeline based methods, sequence-to-sequence (Seq2Seq) based methods become more and more popular recently, because they train agents by mapping from an utterance to another one and require less human efforts and little labeled data~\cite{goal2,goal3}. However, the abundant heterogeneous information may not be fully utilized in Seq2Seq models.
For example, the dialogue act flow information, as illustrated in Figure~\ref{fig:example}, is very important and should be modeled, because this information can reflect the knowledge of act patterns in real human-to-human conversation data and is usually handcrafted in most existing commercial platforms. Intuitively, incorporating heterogeneous information into Seq2Seq based methods could not only reduce the dependence on manual work, but also help to learn better end-to-end dialogue agents.

To this end, this paper investigates the problem of how to model end-to-end dialogue in a data-driven way and leverage heterogeneous dialogue information. To automatically and effectively learn knowledge from data, we resort to representation learning framework~\cite{representation} and propose a joint end-to-end model by multi-task representation learning method to embed multiple information into low-dimensional vectors. 

More specifically, the proposed method contains three main components: (1) Heterogeneous data encoding; (2) Space transformation; and (3) Multi-task prediction. 
In the first component, the proposed method encodes heterogeneous dialogue information into low-dimensional semantic embeddings. Then, the embeddings are transformed from semantic space to a new space, namely \textit{dialogue act flow space}, via an autoencoder in the second component. The learned embeddings in the new space are called DialogAct2Vec, which are expected to capture both the knowledge of semantics and dialogue act flow pattern. In the last component, the DialogAct2Vec are used to simultaneously predict the next act and the next utterance.
The proposed method does not depend on much manual work during system design, and meanwhile it can overcome the limitation of Seq2Seq based methods that cannot fully leverage heterogeneous dialogue information. The model can effectively capture knowledge from multi-task representation learning, which is demonstrated by the experiments on a public dataset for restaurant reservation.

The main contributions of this paper are:


\begin{itemize}
	\item We formulate the problem of dialogue modeling and agent building from the perspective of multi-task representation learning.
	\item We solve the problem by proposing a multi-task representation learning method. The method has several advantages: (1) It can learn knowledge from heterogeneous data; (2) Multi-task learning can greatly improve the effectiveness of learned representations; (3) The method is for end-to-end dialogue modeling and does not require too much human intervention in system design.
	\item Experiments are conducted on a public dataset, and the proposed method outperforms the state-of-the-art baselines on several metrics for both the predicting tasks of act and utterance.
\end{itemize}

\section{Related Work}

There are a lot of research works with end-to-end models for chit-chat conversation~\cite{chat1,chat3}. However, for task-completion dialogues, it is still challenging to develop models without human intervention~\cite{jd}. 
Many existing works treat and learn each module of the pipeline separately, and then concatenate them as one system. For example, a network-based end-to-end trainable system for task-oriented dialogues is proposed~\cite{EACL}, where the each module (i.e. NLU, DST, DC and NLG) is trainable from data. Therefore, the whole system is data-driven without too much handcrafted effort. The trainable end-to-end system can be improved by enhancing the dialogue control module by a recent method~\cite{IJCNLP}. 
However, separately training each module requires a lot of labeled data, and would lead to potential error accumulation along the pipeline~\cite{multitask1}. End-to-end models are studied to learn a mapping from historical utterances to system responses for solving the labeling and error propagation issues~\cite{goal2,goal3}, but it may suffer from lack of good policy control without learning enough knowledge from heterogeneous training data.

Recently, multi-task learning methods are proposed to improve the system's stability because the multiple modules are trained together and the error propagation issue can be alleviated. Moreover, the training efficiency can be improved since multi-task learning would share network parameters and results in less workload for computing. However, integrating all the modules into one multi-task model is still a challenge due to the difficulty of coordinating knowledge representations for different tasks. Only parts of the modules are connected in existing works, such as language understanding and dialogue state tracking~\cite{multitask1} and slot filling in E-commerce domain~\cite{multitask2}. On the other hand, some works propose to model dialogue with knowledgeable representations, but do not involve multiple tasks~\cite{mtrl1}. In this paper, we explore to learn the joint representations of dialogue control and natural language, and expect that the combination of multi-task learning and representation learning could bring about better performance improvement.

Many works based on end-to-end neural networks have been studied for each module of dialogue system, for example, including the slot filling issue in natural language understanding~\cite{nlu1,nlu2}, the dialogue state tracking task~\cite{dst1,dst3,dst4}, the dialogue control task with reinforcement learning~\cite{BBQ,hybrid}, and the natural language generation task~\cite{li,eric-etal-2017-key}. Our work is inspired by those existing neural network methods. A recent work proposes a state tracking framework for multiple domains to collect high-quality training data~\cite{nips}, which shares the similar idea with us but for different motivation, and we both consider to leverage the act sequence information. Another work proposes to combine dialogue self-play and crowd-sourcing to generate fully-annotated dialogues with diverse and natural utterances~\cite{bootstraping}, which shows that end-to-end neural models have great promise towards building conversational agents which can be trained from data. 

\section{Model Architecture}

This paper proposes an end-to-end joint model by multi-task representation learning method, which contains three main components as shown in Figure~\ref{fig:model}: (1) Heterogeneous data encoding; (2) Space transformation; and (3) Multi-task prediction for the next system action and the next system utterance. 

Firstly, we leverage CNN~\cite{cnn2} and BiLSTM~\cite{lstm1} with attention mechanism 
to encode heterogeneous dialogue information, including the act sequence, historical utterances and the current utterance, into embeddings in semantic space. Generally, the assumption of semantic space is that the embeddings should be close if they have some semantic relationship, e.g. the positional relationship in natural language sequence, and otherwise they should be distant~\cite{word2vec}. Similarly, we propose a new \textit{dialog act flow space} which embeds both the semantic knowledge and dialogue flow knowledge, and the embeddings in the space have an assumption that they should be close if they are adjacent (or close) in the dialogue flows and otherwise they are distant. Then, the new embeddings are expected to be superior to predict both the next system act and the next system utterance at the same time. To this end, we propose an autoencoder~\cite{autoencoder} based method in the second component to achieve space transformation from the semantic space to the dialogue act flow space. Finally, the third component performs multi-task prediction, in which the new embeddings are used to predict both tasks simultaneously. We name the learned representations, which can capture the both-side knowledge, with DialogAct2Vec.

\subsection{Problem Description}

\begin{figure*}
	\centering
	\includegraphics[height=2.2in,width=6.4in]{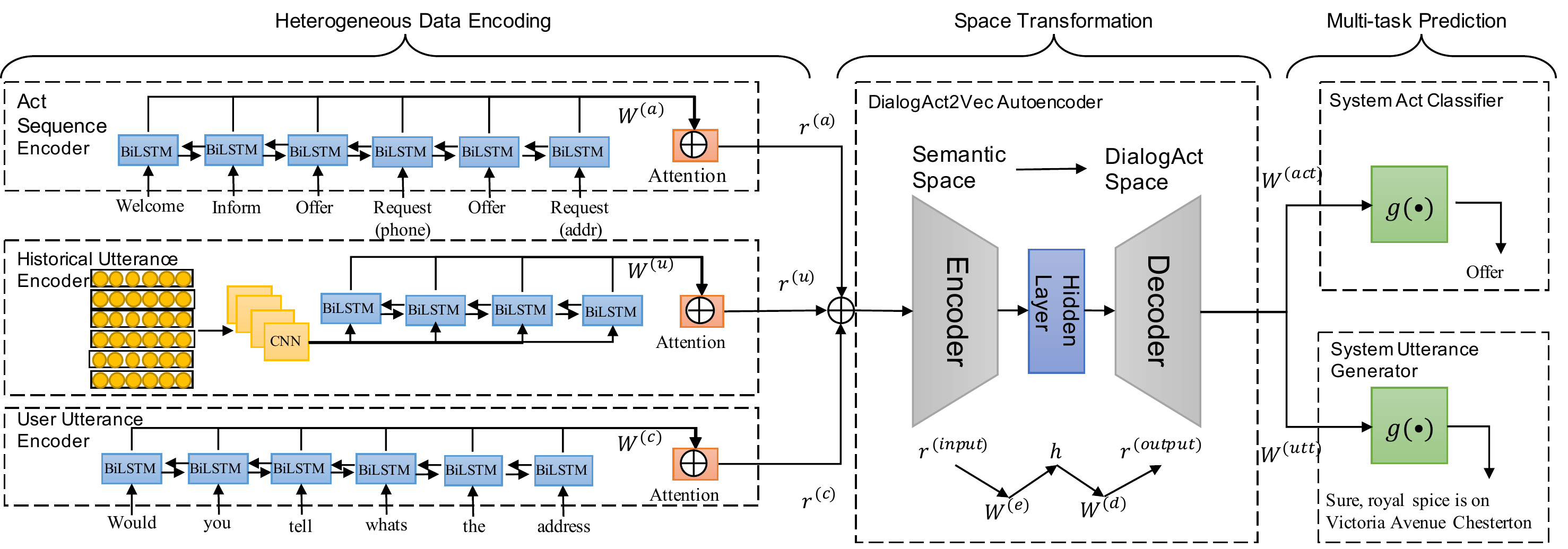}
	\caption{The architecture of our multi-task representation learning model.\label{fig:model}}
\end{figure*}

Assume a dialogue session has $T$ dialogue turns, and we do not distinguish whether it is user's turn or system's turn. For each turn $t\in\{1,...,T\}$, our model takes historical act sequence $\{a_1,...,a_{t-1}\}\in \mathbb{A}$, historical utterances $\{u_1,...,u_{t-1}\}\in \mathbb{U}$ and the current user utterance $u_{t-1}\in \mathbb{U}$ as inputs, and simultaneously outputs the predicted next act $a_t$ and the next utterance $u_t$ which the system agent will say out. As shown in Figure~\ref{fig:model}, our model separately encodes the current user utterance to highlight the user's current states. Our model includes an act sequence encoder, a historical utterance encoder, a user utterance encoder, a DialogAct2Vec autoencoder, a system act classifier and a system utterance generator. In general, the model is designed for better learning the representations in the dialog act flow space from heterogeneous data through multi-task learning, to improve the performance of act prediction and utterance prediction tasks. Next, we will introduce the each component in details.

\subsection{Act Sequence Encoder}

Act information $a_i$ corresponds to each historical utterance $u_i$, where $i\in\{1,...,t-1\}$, and all the acts within a dialogue session are regarded as a sequence in order to utilize the sequential information. We leverage a BiLSTM~\cite{seq2seq} to encode the act sequence information with an attention mechanism which gives the acts different weights for highlighting notable ones. The formal definition is:

\begin{equation}
	h_{i}^{\overrightarrow{(a)}}=LSTM^{\overrightarrow{(a)}}(h_{i-1}^{\overrightarrow{(a)}}, E(a_{i})),
\end{equation}
\begin{equation}
	h_{i}^{\overleftarrow{(a)}}=LSTM^{\overleftarrow{(a)}}(h_{i+1}^{\overleftarrow{(a)}}, E(a_{i})),
\end{equation}
\begin{equation}
	h_{i}^{(a)}=[h_{i}^{\overrightarrow{(a)}},h_{i}^{\overleftarrow{(a)}}],
\end{equation}
\begin{equation}
	e_{i}^{(a)}=tanh(W^{(a)}h_{i}^{(a)})],
\end{equation}
\begin{equation}
	v_{i}^{(a)}=\frac{\exp{(e_i^{(a)})}}{\sum_{i=1}^{t-1}\exp(e_i^{(a)})},
\end{equation}
\begin{equation}
	\textstyle r^{(a)}=\sum_{i=1}^{t-1}{v_{i}^{(a)}}{h_{i}^{(a)}},
\end{equation}
where $h_{i}^{\overrightarrow{(a)}}$ and $h_{i}^{\overleftarrow{(a)}}$ are the hidden states of forward LSTM and backward LSTM, respectively. $E(x)\in\mathbb{R}^k$ means a $k$-dimensional embedding of $x$. Equation 1-3 represent the BiLSTM encoder and Equation 4-6 are for the attention mechanism. $r^{(a)}$ is the final encoded representation of an act sequence.

\subsection{Historical Utterance Encoder}

For a dialogue turn $t$, the data form of historical utterances is a sequence of sentences $\{u_1,...,u_{t-1}\}$, and each sentence $u_i,i\in\{1,...,t-1\}$, consists of a sequence of tokens $\{x_1,...,x_n\}$ where $n$ is the length of the utterance. To capture both the dialogue-level information within sentence sequences and the semantic-level information of token sequences, we preserve the hierarchical structure for modelling and leverage a CNN model 
followed by an attention-based BiLSTM 
to encode the historical utterances.

To get the embedding of an utterance, we concatenate all its token embeddings (padded where necessary) as follows:
\begin{equation}
	E(u_i)=E(x_1)\oplus E(x_2)\oplus...\oplus E(x_n).
\end{equation}
Then, a feature $c_j^i$ of $x_j\in u_i$ for CNN is generated from a window by
\begin{equation}
	c_j^i=tanh(W\cdot E(x_{j:j+h-1})+b),
\end{equation}
where $W\in\mathbb{R}^{h\times k}$ is a filter, $h$ is the window size, $k$ is the token embedding size, $b\in\mathbb{R}$ is a bias term, and $E(x_{j:j+h-1})$ refers to the concatenation of word embeddings from $x_j$ to $x_{j+h-1}$ in $u_i$.

After moving the window from the beginning of an utterance to its end, we can get a feature map: 
\begin{equation}
	c^i=[c_1^i,...,c_{n-h+1}^i]\in\mathbb{R}^{n-h+1}.
\end{equation}
Then, a max-over-time pooling operation over the feature map is performed and the maximum value $\hat{c}^i=\max\{c^i\}$ is taken as the feature corresponding to the filter $W$.

Since multiple filers with different window sizes can be utilized, we can get the representation for each utterance $u_i$ by
\begin{equation}
	r_i^{(u)}=\{\hat{c}_1^i,\hat{c}_2^i,...,\hat{c}_m^i\},
\end{equation}
where $m$ is the number of combinations of different filters and window sizes. The CNN model captures the hierarchical structure information and can also handle variable utterance lengths.

After obtaining the utterance representations $r_i^{(u)}, i\in\{1,...,t-1\}$, similar to the previous act sequence encoder, a BiLSTM with attention mechanism is adopted to capture the semantic information from the sequences, which can be formally described as follows:

\begin{equation}
	h_{i}^{\overrightarrow{(u)}}=LSTM^{\overrightarrow{(u)}}(h_{i-1}^{\overrightarrow{(u)}}, r_i^{(u)}),
\end{equation}
\begin{equation}
	h_{i}^{\overleftarrow{(u)}}=LSTM^{\overleftarrow{(u)}}(h_{i+1}^{\overleftarrow{(u)}}, r_i^{(u)}),
\end{equation}
\begin{equation}
	h_{i}^{(u)}=[h_{i}^{\overrightarrow{(u)}},h_{i}^{\overleftarrow{(u)}}],
\end{equation}
\begin{equation}
	e_{i}^{(u)}=tanh(W^{(u)}h_{i}^{(u)})],
\end{equation}
\begin{equation}
	v_{i}^{(u)}=\frac{\exp{(e_i^{(u)})}}{\sum_{i=1}^{t-1}\exp(e_i^{(u)})},
\end{equation}
\begin{equation}
	\textstyle r^{(u)}=\sum_{i=1}^{t-1}{v_{i}^{(u)}}{h_{i}^{(u)}}.
\end{equation}
$r^{(u)}$ is the final representation for the historical utterances.

\subsection{User Utterance Encoder}

We build a specific encoder for the current user utterance, separately from the previous historical utterances, because the current user's state is expected to be highlighted and learned by this encoder. To encode the utterance $u_{t-1}$ with a token sequence $\{x_1,x_2,...,x_n\}$, we leverage a BiLSTM with attention mechanism. The final representation can be formally described as follows:

\begin{equation}
	h_{i}^{\overrightarrow{(c)}}=LSTM^{\overrightarrow{(c)}}(h_{i-1}^{\overrightarrow{(c)}}, E(x_{i})),
\end{equation}
\begin{equation}
	h_{i}^{\overleftarrow{(c)}}=LSTM^{\overleftarrow{(c)}}(h_{i+1}^{\overleftarrow{(c)}}, E(x_{i})),
\end{equation}
\begin{equation}
	h_{i}^{(c)}=[h_{i}^{\overrightarrow{(c)}},h_{i}^{\overleftarrow{(c)}}],
\end{equation}
\begin{equation}
	e_{i}^{(c)}=tanh(W^{(c)}h_{i}^{(c)})],
\end{equation}
\begin{equation}
	v_{i}^{(c)}=\frac{\exp{(e_i^{(c)})}}{\sum_{i=1}^{n}\exp(e_i^{(c)})},
\end{equation}
\begin{equation}
	\textstyle r^{(c)}=\sum_{i=1}^{n}{v_{i}^{(c)}}{h_{i}^{(c)}},
\end{equation}
where $E(x)$ is the embedding of token $x$ and $r^{(c)}$ is the final output of user utterance encoder.

\subsection{DialogAct2Vec Autoencoder}

To better capture and coordinate the heterogeneous knowledge learned from various sequences of act and utterances, we transform the three learned representations into a common space, namely \textit{Dialog Act Flow Space}. The proposed space is expected to embed both the knowledge of semantics and that of dialogue flows, and we call the embeddings in the new space as DialogAct2Vec. Then, DialogAct2Vec are used to simultaneously predict the next act and the next utterance. The process of space transformation is implemented by an autoencoder, and multi-task learning algorithm is used for the representations to capture the both knowledge.

Specifically, the DialogAct2Vec autoencoder includes three layers: an encoding layer for semantic space, a hidden layer for transformation and a decoding layer for dialog act flow space. The formal definitions are:
\begin{equation}
	r^{(input)}=[r^{(a)}, r^{(u)},r^{(c)}],
\end{equation}
\begin{equation}
	h= W^{(e)}\cdot r^{(input)T},
\end{equation}
\begin{equation}
	r^{(output)}=W^{(d)}\cdot f^{T}(h),
\end{equation}
where $r^{(input)}\in\mathbb{R}^{3k}$ is the concatenated result of previous three representations, $h\in\mathbb{R}^{d}$ is the hidden states, $d$ and $s$ are the sizes of hidden layer and decoding layer, $f(\cdot)$ is an activation function such as sigmoid, $W^{(e)}\in\mathbb{R}^{d\times 3k}$ and $W^{(d)}\in\mathbb{R}^{s\times d}$ are the parameters for encoding and decoding respectively. Note that $r^{(output)}\in\mathbb{R}^{s}$ is namely the transformed representation, DialogAct2Vec, which is then used for the subsequent multi-task prediction.

\subsection{System Act Classifier}

The first task in this work is act prediction. An act classifier is set to connect with the previous DialogAct2Vec autoencoder. The classifier predicts the next act which the system should execute in the next turn $t$, which is a multi-class classification problem.

To leverage multi-task learning, we use a softmax function to calculate each class's probability and choose the class with maximum value as the predicted act. Assume that we have $C^{(a)}$ different classes of acts, the predicted act $\hat{y}^{(a)}$ can be obtained as follows:

\begin{equation}
	\hat{y}^{(a)}=\arg\max_i g_i(W^{(act)}r^{(output)}),
\end{equation}
where $g(\cdot)$ is a softmax function and $W^{(act)}\in\mathbb{R}^{C^{(a)}\times s}$ are the parameters.

\subsection{System Utterance Generator}

The second task in this work is utterance prediction, in which we leverage a retrieval-based generator to produce a system utterance for the next dialogue turn $t$. Intuitively, the KNN-based methods can do well for the retrieval task, but we use the same form with the act classifier in order to make the utterance prediction task be learnable and consistent in our multi-task learning algorithm. This way shares the same goal with the nearest neighbor task, but we change the problem from unsupervised learning to supervised learning.

Assume that there are $C^{(u)}$ classes and each utterance is a class, the predicted target can be obtained as follows:
\begin{equation}
	\hat{y}^{(u)}=\arg\max_i g_i(W^{(utt)}r^{(output)}),
\end{equation}
where $W^{(utt)}\in\mathbb{R}^{C^{(u)}\times s}$ are the parameters to learn.

\subsection{Loss Function for Model Learning}

We leverage a vanilla multi-task learning architecture~\cite{vanilla} and use a cross-entropy loss function for training the model. The two-fold loss function is defined as follows:
\begin{equation}
	\begin{aligned}
		\mathcal{L}=&-\alpha\sum_{i=1}^N\sum_{j=1}^{C^{(a)}} y_j^{(a)}\log [g(W^{(a)}r_i^{(output)})]\\&-(1-\alpha)\sum_{i=1}^N\sum_{j=1}^{C^{(u)}} y_j^{(u)}\log [g(W^{(u)}r_i^{(output)})],
	\end{aligned}
\end{equation}
where $N$ is the number of samples, $y_j^{(a)}$ and $y_j^{(u)}$ are the ground truths (1 or 0) for $C_j^{(a)}$ and $C_j^{(u)}$. The first term is for act prediction and the second is for utterance prediction respectively, and $\alpha$ is a hyperparameter to balance the two tasks.

\section{Experiment}

We report our model's performance over baselines on the two tasks: act prediction and utterance prediction.

\subsection{Evaluation Metrics}

For the act prediction task, since we treat it as a multi-class classification problem with 9 classes, the metrics we employed for evaluation are Micro-F1 and Macro-F1.

For the utterance prediction task, we regard it as retrieval-based utterance generation, which is actually a very sparse multi-class classification problem in our experiment. All the candidate utterances are ranked based on the predicted scores and the top@$k$ ($k=1,3,5,10$) are retrieved for evaluation. We calculate the BLEU@$k$ ($k=1,2,3,4$) and the cumulative BLEU@4 (BLEU@4(cumu)) for evaluating the similarity between the retrieved ones and the ground truths~\cite{bleu}.

Considering that it is an imbalanced multi-class classification problem for act prediction and the utterance prediction task is for generating the right utterance, we focus on Micro-F1 and BLEU@4(cumu) the two metrics for respective tasks.

\subsection{Baselines}

We select recently-proposed methods as our baselines for comparison. Since our tasks are act prediction and utterance prediction, we do not choose those baselines for natural language generation. Also, newer methods for single part, i.e. DC~\cite{BBQ} or NLG~\cite{goal3}, rather than the complete system are not considered in our experiments. The baselines may apply to only one or both of the two tasks. Our model is named with MTRL.

Some variants of our model are also built by removing one or more components from it to see how each component can perform and contribute to the final performance. The compared baselines include:

\begin{itemize}
	\item Seq2Seq with Attention. This is a naive sequence-to-sequence model with attention mechanism that maps an utterance to another one. This model only applies to the utterance prediction task.
	\item BiLSTM-CRF~\cite{baseline1}. This is a specific model for the act prediction task and it is implemented by using BiLSTM and conditional random field (CRF).
	\item HCNs~\cite{hybrid}. This is a recent end-to-end pipeline method learned also from DSTC2 by proposing the Hybrid Code Networks\footnote{\url{http://docs.deeppavlov.ai/en/latest/features/skills/go_bot.html}}. We compare with this model for the utterance prediction task.
	\item HisSeq2Seq (Act or Utt). This model is a Seq2Seq with Attention model with an additional encoder for historical utterances to predict the next act or utterance respectively. This model is for single task.
	\item HisSeq2Seq with Act (Act or Utt). This is HisSeq2Seq model with encoding act sequence information for the act prediction or utterance prediction respectively. This model is also for single task.
	\item SingleMTRL (Act or Utt). This model is a single-task version of our model which utilizes all the heterogeneous data and includes the Autoencoder.
	\item MTRL without Act. This is a multi-task learning model based on the HisSeq2Seq model, i.e. a model by removing the act sequence information from MTRL.
	\item MTRL without Autoencoder. This is a multi-task learning model based on the HisSeq2Seq with Act model, i.e. a model by removing the DialogAct2Vec Autoencoder component from MTRL.
	\item MTRL. This is our proposed end-to-end model by multi-task representation learning method. It models heterogeneous dialogue information and embeds them into low-dimensional vectors by the Autoencoder.
\end{itemize}

\subsection{Dataset}

We use the public dataset DSTC2~\cite{challenge} that belongs to restaurant reservation scenario to evaluate our method and baselines. For each dialogue with $T$ turns, we re-organize the utterance and act data into $T$ samples. For each turn $t=\{1,2,...,T\}$, a sample is created with a historical act sequence and a historical utterance sequence, and the labels are the current act and current utterance. In total, we have 25,437 samples including training set and test set. The statistics of the dataset is listed in Table~\ref{statistic}. We have nine different classes of acts, which is an unbalanced distribution.

\begin{table}
	\begin{center}
		\caption{\label{statistic} Statistics of DSTC2 dataset. }
		\begin{tabular}{|l|c|}
			\hline
			Number of Dialogues & 3,227 \\
			Average Turns per Dialogue & 15.76 \\
			Average Words per Turn & 8.47 \\
			Number of Words & 1,205 \\
			Number of Samples & 25,437 \\
			Number of Tokens & 2,909,852 \\
			\hline
		\end{tabular}
	\end{center}
\end{table}
\begin{table}
	\begin{center}
		\caption{\label{result_act} Performance of act prediction task. }
		\begin{tabular}{|l|c|c|}
			\hline
			Model & Micro-F1 & Macro-F1 \\
			\hline
			BiLSTM-CRF & 0.7381 & 0.3259 \\
			HisSeq2Seq(Act) & 0.6901 & 0.3828 \\
			HisSeq2Seq w/ Act(Act) & 0.7032 & 0.3955 \\
			SingleMTRL(Act) & 0.7159 & 0.4058 \\
			\hline
			MTRL(our model) & \textbf{0.7412} & \textbf{0.4654} \\
			\hline
		\end{tabular}
	\end{center}
\end{table}

\subsection{Training Setting}

We use the stochastic gradient descent (SGD) method 
to optimize our model and set the minibatch size as 32. During training all the baseline models, a subset of 10\% samples are randomly separated from the whole training dataset and they are used for validation. All the experimental results are from 5-fold cross validation.

We adopt the well-trained GloVe embeddings with a dimension size of 300~\cite{glove} as the input token embeddings. If there is a token not existing in the dictionary of GloVe embeddings, zero is used for padding. Although the proportion of unknown tokens is small, only 0.24\% among all the tokens by our statistic, we still tried to use the `unk' token from GloVe dictionary for padding, and find the results are comparable. To reduce the computing workload, we adopt zeros for padding. For the input representations of acts, we build one-hot vectors with a dimension size of the number of different acts. 

For the CNN model built in the historical utterance encoder, we set 3 layers with 32 filers in total and the sizes of filters are 3, 4, 5. The size of hidden state in BiLSTM is set as 80. The size of hidden state in autoencoder is set as 128, and its input size and output size are both 160. The balancing factor $\alpha$ is set as 0.5 in our experiments.

\begin{table*}
	\begin{center}
		\caption{\label{result_utt} Top@3 performance of utterance prediction task. }
		\begin{tabular}{|l|c|c|c|c|c|c|c|}
			\hline
			Model &  BLEU@1 & BLEU@2 & BLEU@3 & BLEU@4 & BLEU@4(cumu) \\
			\hline
			Seq2Seq w/ Attention & 0.2812 & 0.1230 & 0.1631 & 0.2045  & 0.2112 \\
			HCNs & 0.5352 & 0.4915 & 0.4658 & 0.4429 & 0.4488 \\
			HisSeq2Seq(Utt) & 0.5787 & 0.4097 & 0.3444 & 0.3119 & 0.3446 \\
			HisSeq2Seq w/ Act(Utt) & 0.6255 & 0.4771 & 0.4187 & 0.3912 & 0.4178 \\
			SingleMTRL(Utt) & 0.6454 & 0.5049 & 0.4476 & 0.4199 & 0.4468 \\
			\hline
			MTRL(our model) & \textbf{0.6609} & \textbf{0.5237} & \textbf{0.4672} & \textbf{0.4388} & \textbf{0.4672} \\
			\hline
		\end{tabular}
	\end{center}
\end{table*}

\subsection{Results on Act Prediction}

Table~\ref{result_act} shows the performance of multi-class classification for the act prediction task. We can find that both the act sequence information and the DialogAct2Vec Autoencoder are greatly helpful to improve the performance, compared between MTRL and other baselines. Seeing from BiLSTM-CRF's performance, it seems only encoding historical information or act sequence information does not guarantee the best results on Micro-F1. However, combining all the positive components with multi-task learning, like our model, can achieve the state-of-the-art performance, which means the other task of utterance prediction also contributes positively to this task of act prediction.

\begin{table*}
	\begin{center}
		\caption{\label{ablation} Ablation experiment by removing one component from MTRL.}
		\begin{tabular}{|l|c|c|c|c|c|c|c|c|c|}
			\hline
			\multirow{2}{*}{Model} &
			\multicolumn{2}{c|}{\textit{Act Prediction}} &
			\multicolumn{5}{c|}{\textit{Utterance Prediction}} \\
			
			\cline{2-8} & Micro-F1 & Macro-F1 &  BLEU@1 & BLEU@2 & BLEU@3 & BLEU@4 & BLEU@4(cumu) \\
			\hline
			MTRL(our model) & \textbf{0.7412} & \textbf{0.4654} & \textbf{0.6609} & \textbf{0.5237} & \textbf{0.4672} & \textbf{0.4388} & \textbf{0.4672} \\
			MTRL w/o Act & 0.6922 & 0.3665 & 0.5973 & 0.4341 & 0.3703 & 0.3400 & 0.3710 \\
			MTRL w/o Autoencoder & 0.7188 & 0.4123 & 0.6354 & 0.4949 & 0.4376 & 0.4099 & 0.4368 \\
			\hline
		\end{tabular}
	\end{center}
\end{table*}


\subsection{Results on Utterance Prediction}

Table~\ref{result_utt} shows the performance of utterance prediction task by evaluating the top 3 retrieved candidates. Similar to the performance from act prediction (Table~\ref{result_act}), our model can achieve the best performance against all the other baselines in terms of every metric. This also illustrates that the act sequence information, the DialogAct2Vec Autoencoder and multi-task learning can jointly contribute to the performance. Seq2Seq with Attention model uses a generative model for language generation, so the utterance quality is significantly lower than other retrieval based methods. Although the pipeline based HCNs has comparable performance with the SingleMTRL(Utt) model in terms of BLEU@4(cumu), the later model requires much less handcrafted work during system design. This experiment indirectly suggests that the task of predicting the next act can also help to better predict the next utterance.

\begin{figure*}[t!]
	\begin{minipage}[t]{0.33\linewidth}
		\centering
		\includegraphics[width=1\textwidth]{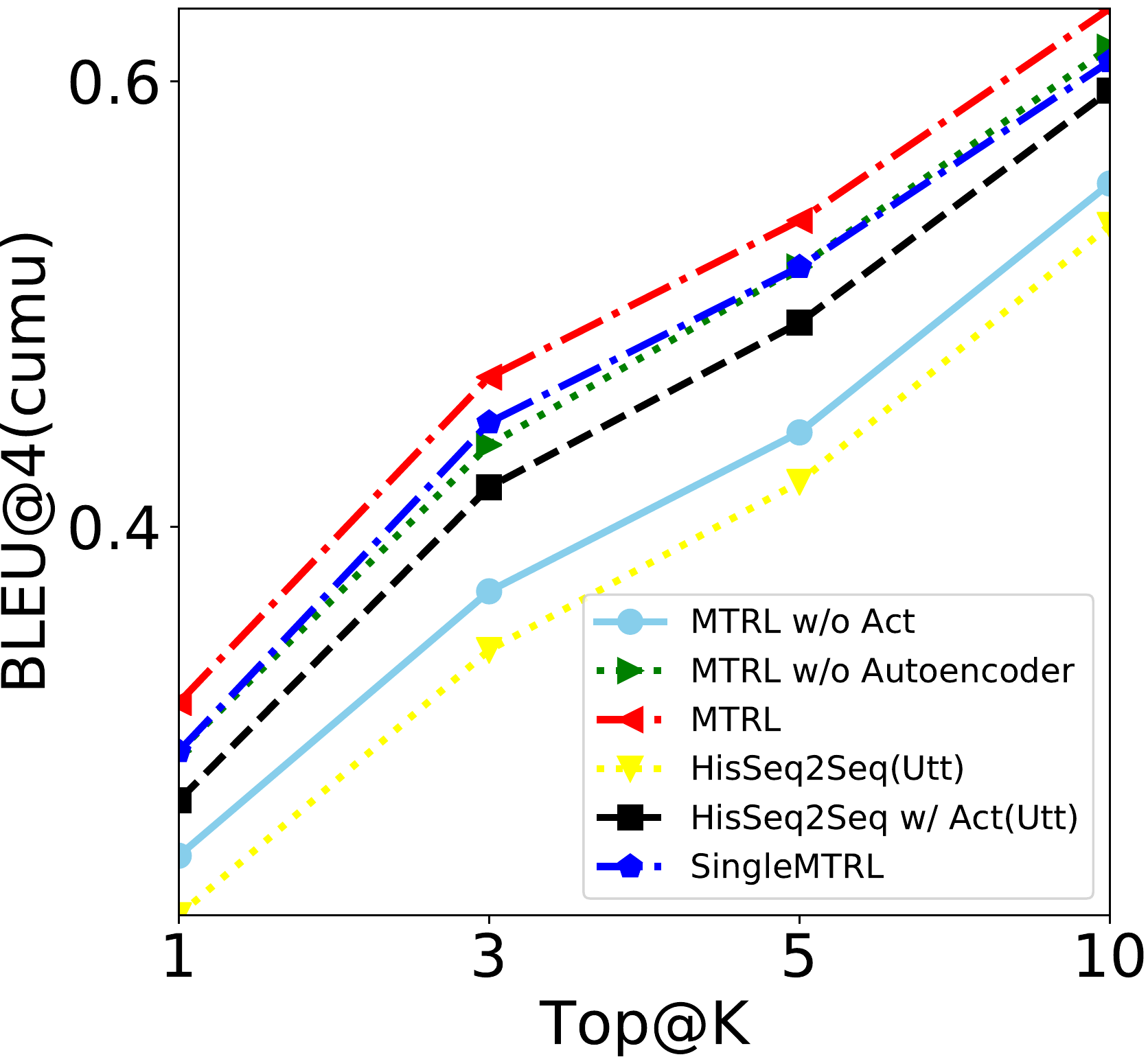}\\
		\caption{Top@$k$ performance for utterance prediction task.\label{fig:top}}
		
	\end{minipage}
	\hfill
	\begin{minipage}[t]{0.66\linewidth}
		\centering
		\includegraphics[width=.49\textwidth]{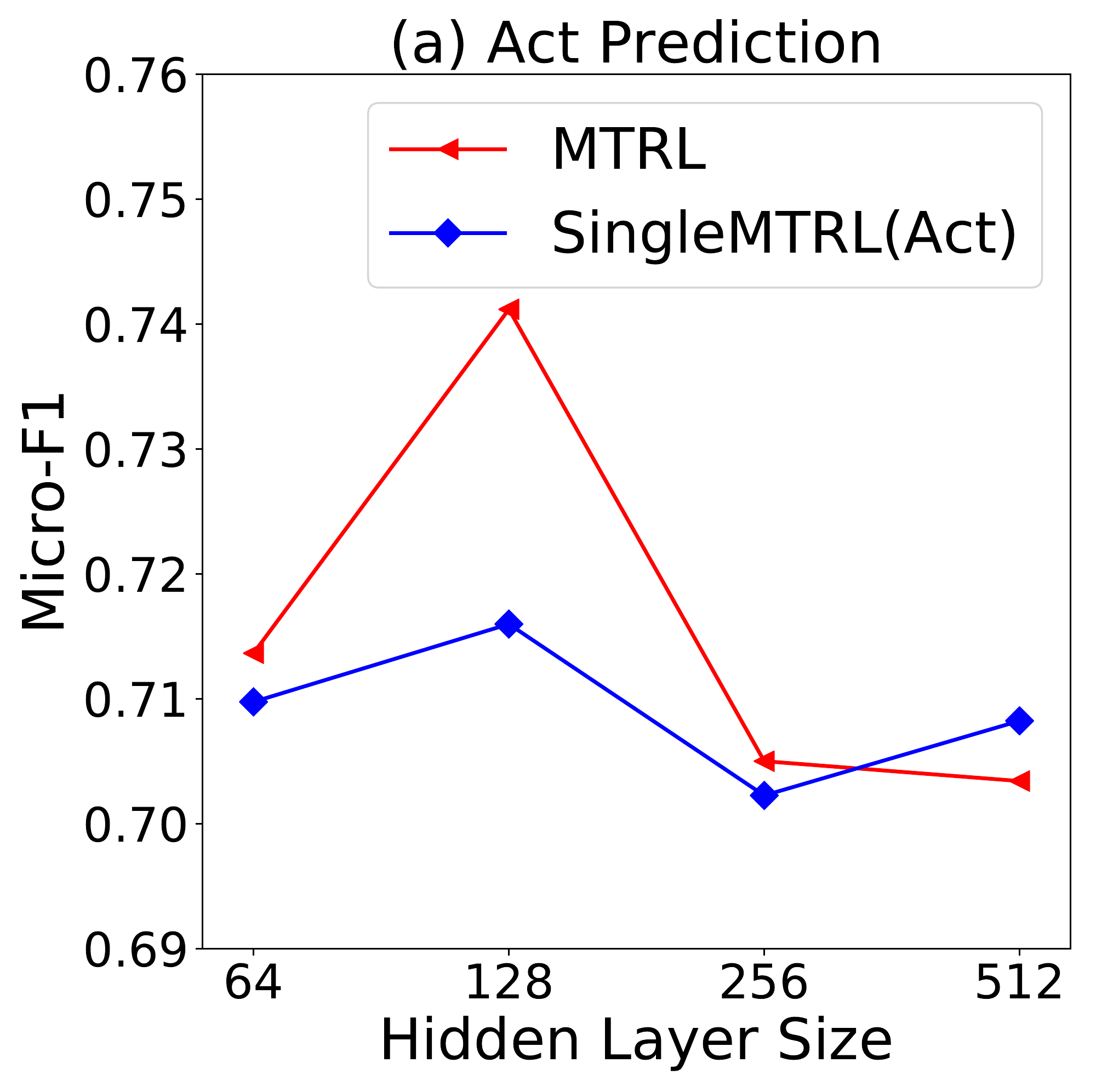}\hfill
		\includegraphics[width=.49\textwidth]{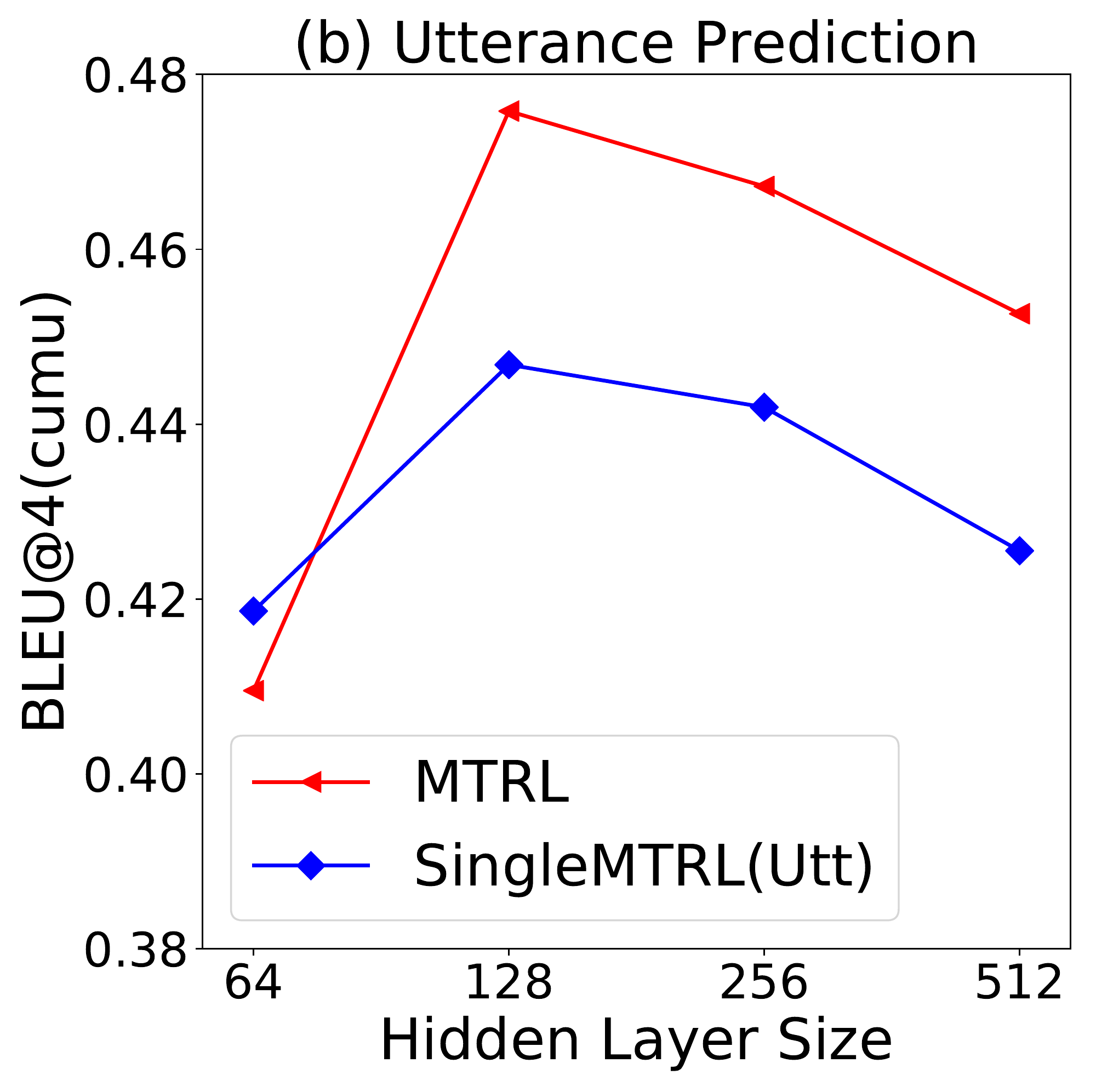}
		\caption{Performance comparison with different hidden layer sizes in DialogAct2Vec Autoencoder.~\label{fig:dim}}
	\end{minipage}
\end{figure*}

We also evaluate the performance with different numbers of retrieved candidates in terms of BLEU@4(cumu) metric. Figure~\ref{fig:top} shows the results by setting the number $k$ as 1, 3, 5 and 10 respectively.
We find that the curves are changing consistently, which illustrates that our model is stable.

\begin{figure}
	\centering
	\includegraphics[width=0.49\textwidth]{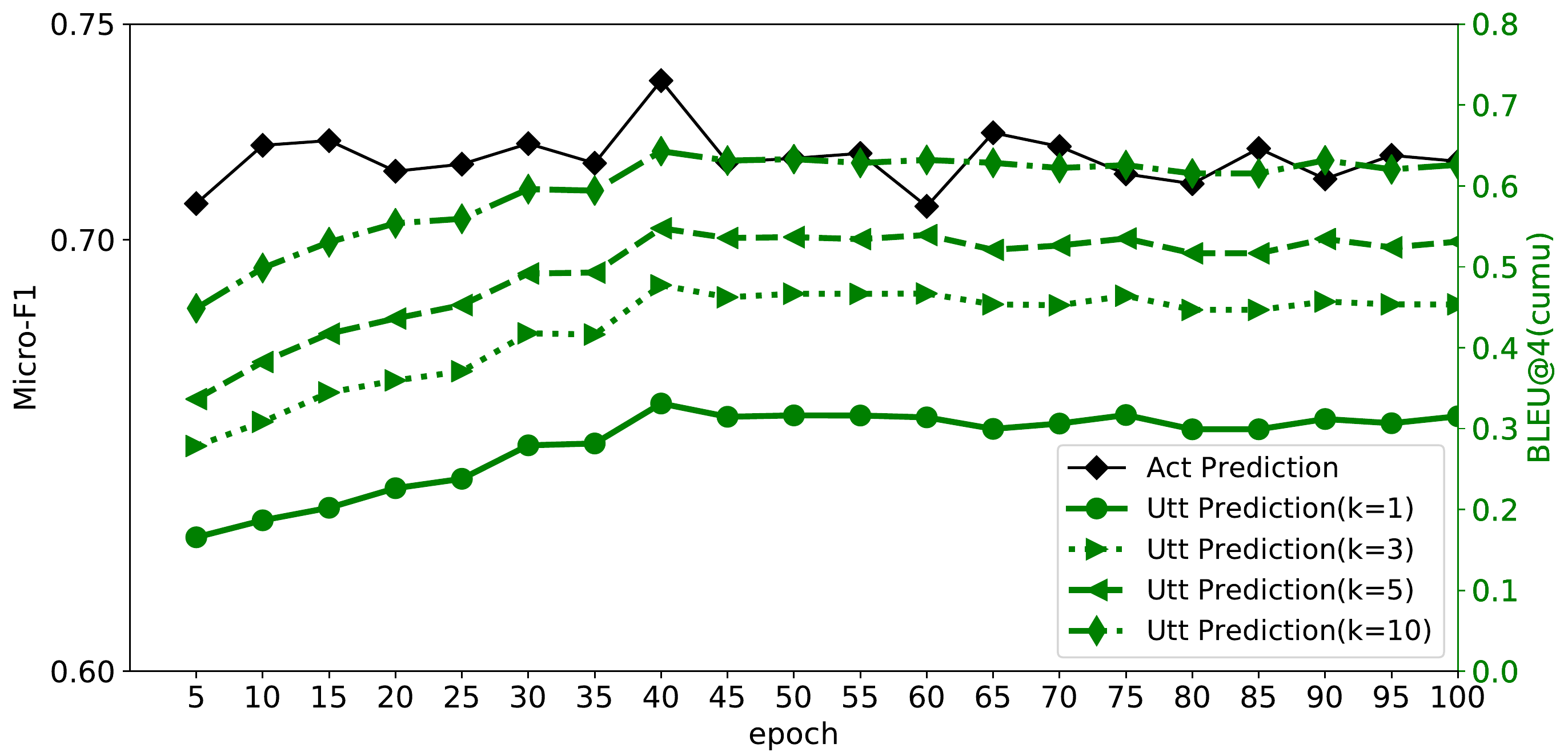}
	\caption{Training curve of our model.\label{fig:training}}
\end{figure}

\subsection{Ablation}

To further analyze the reason why our model can achieve better performance, we conduct several ablation experiments in Table 4 to evaluate the different components in our model. By removing the act sequence information for encoding, both the tasks are affected greatly. It means that (1) the act information is important and (2) our model captures the knowledge well. The similar finding is reflected by MTRL w/o Autoencoder. The phenomenon that the MTRL w/o Act model is worse than the MTRL w/o Autoencoder model is consistent, which suggests that the act information modeling is more significant than the space transformation. However, despite that, the transformation is still necessary.

\subsection{Hyperparameters Sensitivity}

Here, we explore various hyperparameter settings to evaluate our method. Figure~\ref{fig:training} shows the training curve of our model on Micro-F1 and BLEU@4(cumu) for the act prediction task (left $y$ axis) and the utterance prediction task (right $y$ axis) respectively. We can see that, with increasing number of epochs, our model can obtain better performance, and after about 40 epochs of training, the model becomes stable.


To evaluate the capability of the proposed DialogAct2Vec Autoencoder component, we compare the models with different hidden layer sizes, ranging from 64, 128, 256 to 512. Figure~\ref{fig:dim}(a) shows the results of act prediction task and Figure~\ref{fig:dim}(b) is for the utterance prediction task. We can find that setting the size as 128 can achieve optimal performance.


\section{Conclusion and Future Work}

To avoid too much manual intervention and fully utilize various information in end-to-end dialogue agent modeling, this paper proposes a novel end-to-end model by multi-task representation learning. The proposed method is scalable and effective to capture heterogeneous information, and features translating semantic embeddings to a new \textit{dialog act flow space} through a DialogAct2Vec Autoencoder. Experiments demonstrate that (1) the newly-learned representations can well capture both the semantic knowledge and dialogue act flow knowledge, and (2) multi-task learning is helpful to learn better representations for both the tasks of act prediction and utterance prediction. 
In the future, we will explore how to integrate generative models in end-to-end dialogue agent learning, which can generate more diverse utterances.

\small
\bibliography{bibfile}

\begin{thebibliography}{}

\bibitem[\protect\citeauthoryear{Arora \bgroup et al\mbox.\egroup
  }{2018}]{vanilla}
Arora, H.; Kumar, R.; Krone, J.; and Li, C.
\newblock 2018.
\newblock Multi-task learning for continuous control.
\newblock arXiv preprint arXiv:1802.01034.

\bibitem[\protect\citeauthoryear{Bapna \bgroup et al\mbox.\egroup
  }{2017}]{nlu1}
Bapna, A.; T\"{u}r, G.; Hakkani-T\"{u}r, D.; and Heck, L.
\newblock 2017.
\newblock Sequential dialogue context modeling for spoken language
  understanding.
\newblock In {\em SIGDIAL},  103--114.

\bibitem[\protect\citeauthoryear{Bengio, Courville, and
  Vincent}{2013}]{representation}
Bengio, Y.; Courville, A.; and Vincent, P.
\newblock 2013.
\newblock Representation learning: A review and new perspectives.
\newblock {\em IEEE Transactions on Pattern Analysis and Machine Intelligence}
  35(8):1798--1828.

\bibitem[\protect\citeauthoryear{Bordes, Boureau, and Weston}{2017}]{goal1}
Bordes, A.; Boureau, Y.-L.; and Weston, J.
\newblock 2017.
\newblock Learning end-to-end goal-oriented dialog.
\newblock In {\em ICLR}.

\bibitem[\protect\citeauthoryear{Cai \bgroup et al\mbox.\egroup }{2017}]{icdm}
Cai, R.; Zhu, B.; Liu, W.; Ji, L.; Yan, J.; and Hao, T.
\newblock 2017.
\newblock An cnn-lstm attention approach to understanding user query intent
  from online health communities.
\newblock In {\em ICDM Workshops},  430--437.

\bibitem[\protect\citeauthoryear{Chen \bgroup et al\mbox.\egroup }{2017}]{jd}
Chen, H.; Liu, X.; Yin, D.; and Tang, J.
\newblock 2017.
\newblock A survey on dialogue systems- recent advances and new frontiers.
\newblock {\em ACM SIGKDD Explorations Newsletter} 19(2):25--35.

\bibitem[\protect\citeauthoryear{Dhingra \bgroup et al\mbox.\egroup
  }{2017}]{movie}
Dhingra, B.; Li, L.; Li, X.; Gao, J.; Chen, Y.-N.; Ahmed, F.; and Deng, L.
\newblock 2017.
\newblock Towards end-to-end reinforcement learning of dialogue agents for
  information access.
\newblock In {\em ACL},  484--495.

\bibitem[\protect\citeauthoryear{Eric \bgroup et al\mbox.\egroup
  }{2017}]{eric-etal-2017-key}
Eric, M.; Krishnan, L.; Charette, F.; and Manning, C.~D.
\newblock 2017.
\newblock Key-value retrieval networks for task-oriented dialogue.
\newblock In {\em SIGDIAL},  37--49.

\bibitem[\protect\citeauthoryear{Goel \bgroup et al\mbox.\egroup }{2018}]{nips}
Goel, R.; Paul, S.; Chung, T.; Lecomte, J.; Mandal, A.; and Hakkani-Tur, D.
\newblock 2018.
\newblock Flexible and scalable state tracking framework for goal-oriented
  dialogue systems.
\newblock In {\em NeurIPS}.

\bibitem[\protect\citeauthoryear{Gong \bgroup et al\mbox.\egroup
  }{2019}]{multitask2}
Gong, Y.; Luo, X.; Zhu, Y.; Ou, W.; Li, Z.; Zhu, M.; Zhu, K.~Q.; Duan, L.; and
  Chen1, X.
\newblock 2019.
\newblock Deep cascade multi-task learning for slot filling in online shopping
  assistant.
\newblock In {\em AAAI}.

\bibitem[\protect\citeauthoryear{Henderson, Thomson, and
  Williams}{2014a}]{challenge}
Henderson, M.; Thomson, B.; and Williams, J.
\newblock 2014a.
\newblock The second dialog state tracking challenge.
\newblock In {\em SIGDIAL},  263--272.

\bibitem[\protect\citeauthoryear{Henderson, Thomson, and
  Williams}{2014b}]{dstc3}
Henderson, M.; Thomson, B.; and Williams, J.~D.
\newblock 2014b.
\newblock The third dialog state tracking challenge.
\newblock In {\em SLT},  324--329.

\bibitem[\protect\citeauthoryear{Joshi, Mi, and Faltings}{2017}]{goal2}
Joshi, C.~K.; Mi, F.; and Faltings, B.
\newblock 2017.
\newblock Personalization in goal-oriented dialog.
\newblock In {\em NIPS}.

\bibitem[\protect\citeauthoryear{Kalchbrenner, Grefenstette, and
  Blunsom}{2014}]{cnn2}
Kalchbrenner, N.; Grefenstette, E.; and Blunsom, P.
\newblock 2014.
\newblock A convolutional neural network for modelling sentences.
\newblock In {\em ACL},  655--665.

\bibitem[\protect\citeauthoryear{Kumar \bgroup et al\mbox.\egroup
  }{2018}]{baseline1}
Kumar, H.; Agarwal, A.; Dasgupta, R.; and Joshi, S.
\newblock 2018.
\newblock Dialogue act sequence labeling using hierarchical encoder with crf.
\newblock In {\em AAAI},  3440--3447.

\bibitem[\protect\citeauthoryear{Li \bgroup et al\mbox.\egroup }{2016}]{li}
Li, J.; Monroe, W.; Ritter, A.; Galley, M.; Gao, J.; and Jurafsky, D.
\newblock 2016.
\newblock Deep reinforcement learning for dialogue generation.
\newblock In {\em EMNLP},  1192--1202.

\bibitem[\protect\citeauthoryear{Li \bgroup et al\mbox.\egroup }{2017}]{IJCNLP}
Li, X.; Chen, Y.-N.; Li, L.; Gao, J.; and Celikyilmaz, A.
\newblock 2017.
\newblock End-to-end task-completion neural dialogue systems.
\newblock In {\em IJCNLP},  733--743.

\bibitem[\protect\citeauthoryear{Lipton \bgroup et al\mbox.\egroup
  }{2018}]{BBQ}
Lipton, Z.; Li, X.; Gao, J.; Li, L.; Ahmed, F.; and Deng, L.
\newblock 2018.
\newblock Bbq-networks: Efficient exploration in deep reinforcement learning
  for task-oriented dialogue systems.
\newblock In {\em AAAI},  5237--5244.

\bibitem[\protect\citeauthoryear{Liu and Lane}{2017}]{errorpro2}
Liu, B., and Lane, I.
\newblock 2017.
\newblock An end-to-end trainable neural network model with belief tracking for
  task-oriented dialog.
\newblock In {\em INTERSPEECH},  2506--2510.

\bibitem[\protect\citeauthoryear{Liu \bgroup et al\mbox.\egroup
  }{2018}]{withhuman}
Liu, B.; T\"{u}r, G.; Hakkani-T\"{u}r, D.; Shah, P.; and Heck, L.
\newblock 2018.
\newblock Dialogue learning with human teaching and feedback in end-to-end
  trainable task-oriented dialogue systems.
\newblock In {\em NAACL},  2060--2069.

\bibitem[\protect\citeauthoryear{Luo \bgroup et al\mbox.\egroup }{2019}]{goal3}
Luo, L.; Huang, W.; Zeng, Q.; Nie, Z.; and Sun, X.
\newblock 2019.
\newblock Learning personalized end-to-end goal-oriented dialog.
\newblock In {\em AAAI}.

\bibitem[\protect\citeauthoryear{Ma and Hovy}{2016}]{crflstm}
Ma, X., and Hovy, E.
\newblock 2016.
\newblock End-to-end sequence labeling via bi-directional lstm-cnns-crf.
\newblock In {\em ACL},  1064--1074.

\bibitem[\protect\citeauthoryear{Mikolov \bgroup et al\mbox.\egroup
  }{2013}]{word2vec}
Mikolov, T.; Sutskever, I.; Chen, K.; Corrado, G.~S.; and Dean, J.
\newblock 2013.
\newblock Distributed representations of words and phrases and their
  compositionality.
\newblock In {\em NIPS},  3111--3119.

\bibitem[\protect\citeauthoryear{Mrk{\v{s}}i{\'c} \bgroup et al\mbox.\egroup
  }{2017}]{dst1}
Mrk{\v{s}}i{\'c}, N.; S{\'e}aghdha, D.~{\'O}.; Wen, T.-H.; Thomson, B.; and
  Young, S.
\newblock 2017.
\newblock Neural belief tracker: Data-driven dialogue state tracking.
\newblock In {\em ACL},  1777--1788.

\bibitem[\protect\citeauthoryear{Palangi \bgroup et al\mbox.\egroup
  }{2016}]{lstm1}
Palangi, H.; Deng, L.; Shen, Y.; Gao, J.; He, X.; Chen, J.; Song, X.; and Ward,
  R.
\newblock 2016.
\newblock Deep sentence embedding using long short-term memory networks:
  analysis and application to information retrieval.
\newblock {\em TASLP} 24(4):694--707.

\bibitem[\protect\citeauthoryear{Papineni \bgroup et al\mbox.\egroup
  }{2002}]{bleu}
Papineni, K.; Roukos, S.; Ward, T.; and Zhu, W.-J.
\newblock 2002.
\newblock Bleu: a method for automatic evaluation of machine translation.
\newblock In {\em ACL},  311--318.

\bibitem[\protect\citeauthoryear{Peng \bgroup et al\mbox.\egroup
  }{2017}]{hierarchical}
Peng, B.; Li, X.; Li, L.; Gao, J.; Celikyilmaz, A.; Lee, S.; and Wong, K.-F.
\newblock 2017.
\newblock Composite task-completion dialogue policy learning via hierarchical
  deep reinforcement learning.
\newblock In {\em EMNLP},  2231--2240.

\bibitem[\protect\citeauthoryear{Pennington, Socher, and Manning}{2014}]{glove}
Pennington, J.; Socher, R.; and Manning, C.~D.
\newblock 2014.
\newblock Glove: Global vectors for word representation.
\newblock In {\em EMNLP},  1532--1543.

\bibitem[\protect\citeauthoryear{Rastogi, Gupta, and
  Hakkani-Tur}{2018}]{multitask1}
Rastogi, A.; Gupta, R.; and Hakkani-Tur, D.
\newblock 2018.
\newblock Multi-task learning for joint language understanding and dialogue
  state tracking.
\newblock In {\em SIGDIAL},  376--384.

\bibitem[\protect\citeauthoryear{Ren \bgroup et al\mbox.\egroup
  }{2018}]{bilstmcrf1}
Ren, L.; Xie, K.; Chen, L.; and Yu, K.
\newblock 2018.
\newblock Towards universal dialogue state tracking.
\newblock In {\em EMNLP},  2780--2786.

\bibitem[\protect\citeauthoryear{Ritter, Cherry, and Dolan}{2011}]{chat1}
Ritter, A.; Cherry, C.; and Dolan, W.~B.
\newblock 2011.
\newblock Data-driven response generation in social media.
\newblock In {\em EMNLP},  583--593.

\bibitem[\protect\citeauthoryear{Shah \bgroup et al\mbox.\egroup
  }{2018}]{bootstraping}
Shah, P.; Hakkani-T\"{u}r, D.; Liu, B.; and T\"{u}r, G.
\newblock 2018.
\newblock Bootstrapping a neural conversational agent with dialogue self-play,
  crowdsourcing and on-line reinforcement learning.
\newblock In {\em NAACL},  41--51.

\bibitem[\protect\citeauthoryear{Socher \bgroup et al\mbox.\egroup
  }{2011}]{autoencoder}
Socher, R.; Lin, C. C.-Y.; Ng, A.~Y.; and Manning, C.~D.
\newblock 2011.
\newblock Parsing natural scenes and natural language with recursive neural
  networks.
\newblock In {\em ICML},  129--136.

\bibitem[\protect\citeauthoryear{Su, Yuan, and Chen}{2018}]{nlu2}
Su, S.-Y.; Yuan, P.-C.; and Chen, Y.-N.
\newblock 2018.
\newblock How time matters: Learning time-decay attention for contextual spoken
  language understanding in dialogues.
\newblock In {\em NAACL},  2133--2142.

\bibitem[\protect\citeauthoryear{Sutskever, Vinyals, and Le}{2014}]{seq2seq}
Sutskever, I.; Vinyals, O.; and Le, Q.~V.
\newblock 2014.
\newblock Sequence to sequence learning with neural networks.
\newblock In {\em NIPS},  3104--3112.

\bibitem[\protect\citeauthoryear{Wen \bgroup et al\mbox.\egroup }{2017}]{EACL}
Wen, T.-H.; Vandyke, D.; Mrk\v{s}i\'{c}, N.; Ga\v{s}i\'{c}, M.; Rojas-Barahona,
  L.~M.; Su, P.-H.; Ultes, S.; and Young, S.
\newblock 2017.
\newblock A network-based end-to-end trainable task-oriented dialogue system.
\newblock In {\em EACL},  438--449.

\bibitem[\protect\citeauthoryear{Wen \bgroup et al\mbox.\egroup }{2018}]{mtrl1}
Wen, H.; Liu, Y.; Che, W.; Qin, L.; and Liu, T.
\newblock 2018.
\newblock Sequence-to-sequence learning for task-oriented dialogue with
  dialogue state representation.
\newblock In {\em COLING},  3781--3792.

\bibitem[\protect\citeauthoryear{Williams and Zweig}{2016}]{kefu}
Williams, J.~D., and Zweig, G.
\newblock 2016.
\newblock End-to-end lstm-based dialog control optimized with supervised and
  reinforcement learning.
\newblock arXiv:1606.01269.

\bibitem[\protect\citeauthoryear{Williams, Asadi, and Zweig}{2017}]{hybrid}
Williams, J.~D.; Asadi, K.; and Zweig, G.
\newblock 2017.
\newblock Hybrid code networks: practical and efficient end-to-end dialog
  control with supervised and reinforcement learning.
\newblock In {\em ACL},  665--677.

\bibitem[\protect\citeauthoryear{Xu and Hu}{2018}]{dst4}
Xu, P., and Hu, Q.
\newblock 2018.
\newblock An end-to-end approach for handling unknown slot values in dialogue
  state tracking.
\newblock In {\em ACL},  1448--1457.

\bibitem[\protect\citeauthoryear{Zhang \bgroup et al\mbox.\egroup
  }{2018}]{chat3}
Zhang, W.-N.; Cui, Y.; Wang, Y.; Zhu, Q.; Li, L.; Zhou, L.; and Liu, T.
\newblock 2018.
\newblock Context-sensitive generation of open-domain conversational responses.
\newblock In {\em COLING},  2437--2447.

\bibitem[\protect\citeauthoryear{Zhong, Xiong, and Socher}{2018}]{dst3}
Zhong, V.; Xiong, C.; and Socher, R.
\newblock 2018.
\newblock Global-locally self-attentive dialogue state tracker.
\newblock In {\em ACL},  1458--1467.

\end{thebibliography}
\bibliographystyle{aaai}
\end{document}